\documentclass[conference,a4paper]{IEEEtran}
\usepackage{amsmath,graphicx}
\usepackage{times}
\usepackage{epsfig}
\usepackage{amssymb}
\usepackage{nicefrac}
\usepackage{epstopdf}
\usepackage{booktabs}
\usepackage{multirow}
\usepackage[table]{xcolor}
\usepackage{caption}
\usepackage{subcaption}
\usepackage{dblfloatfix}
\usepackage[linesnumbered,ruled]{algorithm2e}
\usepackage{algpseudocode}
\newcommand{\eg}{e.g.}
\newcommand{\ie}{i.e.}
\usepackage{bm}
\usepackage[pagebackref=true,breaklinks=true,colorlinks,bookmarks=false]{hyperref}
\begin{document}
    
    \title{Predicting Privileged Information \\for Height Estimation}
    
    \author{\IEEEauthorblockN{Nikolaos Sarafianos$^{1}$ \qquad Christophoros  Nikou$^{1,2}$  \qquad Ioannis A. Kakadiaris$^{1}$}\\
        
        \IEEEauthorblockN{$^{1}$ Computational Biomedicine Lab, Department of Computer Science, University of Houston \\
            $^{2}$Department of Computer Science and Engineering, University of Ioannina, Ioannina, Greece}
        \IEEEauthorblockA{nsarafianos@uh.edu, cnikou@cs.uoi.gr, ioannisk@uh.edu}
    }    
    \maketitle
    
    \begin{abstract}
        In this paper, we propose a novel regression-based method for employing privileged information to estimate the height using human metrology. The actual values of the anthropometric measurements are difficult to estimate accurately using state-of-the-art computer vision algorithms. Hence, we use ratios of anthropometric measurements as features. Since many anthropometric measurements are not available at test time in real-life scenarios, we employ a learning using privileged information (LUPI) framework in a regression setup. Instead of using the LUPI paradigm for regression in its original form (\ie, \(\epsilon\)-SVR+), we train regression models that predict the privileged information at test time. The predictions are then used, along with observable features, to perform height estimation. Once the height is estimated, a mapping to classes is performed. We demonstrate that the proposed approach can estimate the height better and faster than the \(\epsilon\)-SVR+ algorithm and report results for different genders and quartiles of humans.  
    \end{abstract}
    
    \section{Introduction}\label{sec:Intro}
    Estimating the height of a human can be a challenging task under certain conditions. For example, a robbery suspect description sent by the University of Houston Police Department in April 2016 was the following: ``\textit{unknown race male, \textbf{approximately} 5'02"-5'04" tall, red and white shirt, blue jean pants, and a red hat}''. A question that arises from reading this description is that if humans struggle at estimating the height of another person, then what about computers? In this work, we investigate the challenging task of estimating the height using human metrology.
    
    Integrating soft biometrics such as gender, height, weight, age, and ethnicity to a primary biometrics system (\eg, face) has been studied by Jain \textit{et al.}~\cite{Jain_2004_14918}. In most of the existing literature \cite{victor2002evaluation,Makinen_2008_8474}, the problem of human classification assisted by soft biometrics has been approached using facial information. However, in real-life scenarios, such information might not be available (\eg, the face might be covered or occluded). This led to methods that employ information from the human body to perform human identification and tracking based on soft biometrics \cite{williams2010body,islam2014preliminary, chen2014soft}. Adjeroh \textit{et al.}~\cite{D_2010_11241} studied the correlation of several anthropometric measurements from the CAESAR anthropometric database \cite{Caesar} and proposed a cluster-driven prediction model which employs information from human metrology. In the work of Guo \textit{et al.}~\cite{Cao_2012_16764} the same dataset was used and a method that predicts the gender and the weight was proposed. 
    \begin{figure}[t] 
        \centering
        \includegraphics[width=0.4\textwidth]{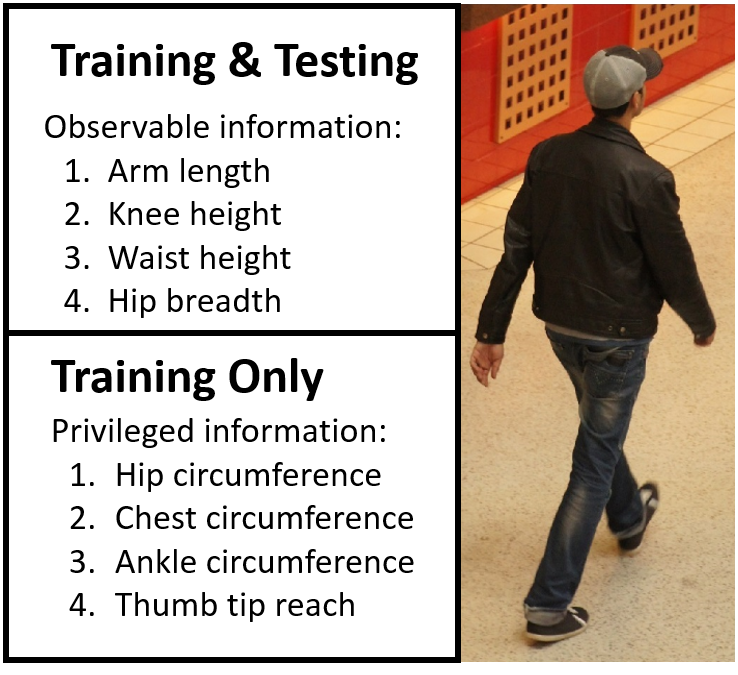}
        \caption{Example of the two types of anthropometric features used throughout this work. Privileged information is employed at training time and is not available during testing.}
        \label{fig:Priv}
    \end{figure}
    
    In 2009, the learning using privileged information (LUPI) framework was introduced by Vapnik and Vashist \cite{Vapnik_2009_15070}. This new paradigm places a nontrivial teacher who provides additional information (\ie, features) during the training process, but it is not available for test examples. It can be applied to both classification (\ie, SVM+ algorithm) and regression tasks (\ie, \(\epsilon\)-SVR+ algorithm). Following this work, new approaches  that leverage privileged information in different ways have been introduced. In the work of Sharmanska \textit{et al.}~\cite{Sharmanska_2014_16759}, samples are examined whether they are easy or difficult to classify in the privileged space. This information (\ie, distance from the margin) is then transferred to the observable space to improve the prediction performance. Lapin \textit{et al.}~\cite{Lapin_2014_16384} related the privileged information framework to the importance of sample weighting and showed that prior knowledge can be encoded using weights in a regular support vector machine. Recently, the LUPI paradigm was employed with applications on biometrics \cite{Xu_2015_16245,Wang_2015_16756,kakadiaris2016body} such as face verification, person identification, age estimation, and gender classification. 
    \begin{figure*}[t] 
        \centering
        \includegraphics[width=0.98\textwidth]{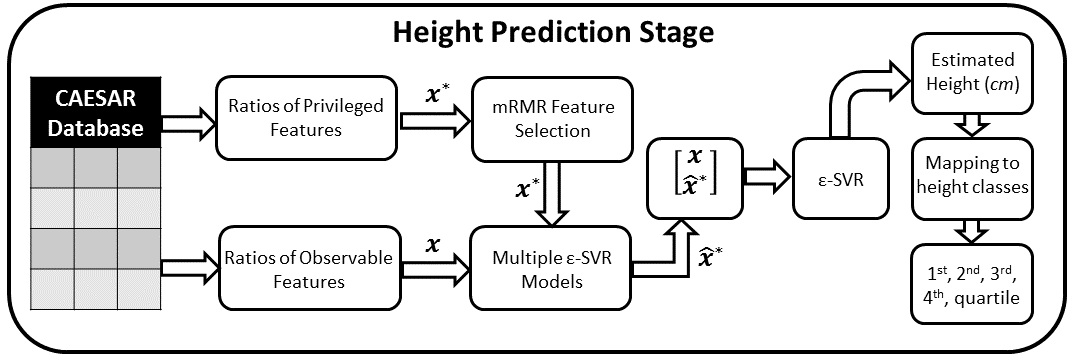}
        \caption{Framework of the proposed approach for height prediction. Given an anthropometric database \cite{Caesar}, ratios of anthropometric measurements are computed for observable and privileged features. For the vector of privileged features \(x^*\), feature selection is performed and the \(K\) most informative features are obtained. Then, \(K\) \(\epsilon\)-SVR models are trained, one for each feature, to predict \(x^*\) from \(x\). This predicted vector is represented by \(\hat{x}^*\). Then, \(\hat{x}^*\) is concatenated with \(x\), and a single \(\epsilon\)-SVR model is learned to predict the height and a mapping to height classes is performed.}
        \label{fig:Method}
    \end{figure*}
    
    We propose  a novel method to estimate the height of a human using ratios of anthropometric measurements by estimating privileged information at test time. Figure~\ref{fig:Priv} depicts some examples of anthropometric measurements which are considered privileged or observable depending on their ability to be estimated from state-of-the-art computer vision algorithms. In this work, we treat as observable information all the measurements that can be obtained from an image when a 3D pose estimation algorithm is used which estimates the human's skeleton. Any other type of information is considered to be privileged and will not be available at test time. Given as an input, observable and privileged anthropometric measurements, their ratios are first computed in each case (\eg, arm length divided by knee height). Then, a feature selection is performed in the privileged space to obtain the most informative privileged features for the height prediction task. Next, we train multiple support vector regression (\(\epsilon\)-SVR) models that predict the privileged information from the observable space at test time. We concatenate the predictions with the initial observable features and train an additional \(\epsilon\)-SVR model which predicts the height from the updated feature vector. Finally, a mapping to height classes is performed. The key difference of the proposed approach with \(\epsilon\)-SVR+ is that privileged information is predicted at test time and used accordingly to obtain a smaller regression error. In \(\epsilon\)-SVR+, the privileged information is employed to further constrain the optimization solution and obtain a more accurate result in the observable space. An overview of our method is depicted in Figure~\ref{fig:Method}. 
    
    The rest of the paper is organized as follows. In Section~\ref{sec:Method}, (i) the ratios of anthropometric measurements are described, (ii) the \(\epsilon\)-SVR+ algorithm proposed by Vapnik and Vashist \cite{Vapnik_2009_15070} is presented, and (iii) our method of privileged information prediction at test time is introduced. In Section~\ref{sec:Exp}, our experimental evaluation is presented and in Section~\ref{sec:Outro}, our work is concluded with a discussion of our findings.
    
    \section{Privileged Information for Height Estimation}\label{sec:Method}
    
    In this section, we describe the LUPI paradigm \cite{Vapnik_2009_15070} for biometrics using human metrology and present the notation used throughout this paper. Then, we introduce the features used to predict the height and present the \(\epsilon\)-SVR+ algorithm \cite{Vapnik_2009_15070} and its implementation. Finally, we introduce the proposed \textit{Privileged Information Prediction} (PIP) method. 
    
    We will refer to \(\bm{x}\) and \(\bm{x}^*\) as the vectors of observable and privileged features, respectively, whereas \(\cal{X}\) and \(\cal{X}^*\) will be denoted as observable and privileged spaces. In the support vector regression setting, a set of \(l\) training data \((x_1,y_1),\dots, (x_l,y_l)\) is provided, where \(\bm{x} \in \cal{X}\) represents a feature vector and \(\bm{y} \in (-\infty, \infty)\) is a real value we want to predict using a regression function \(y = f(x)\). When privileged information is employed, we are also given \(\bm{x}^* \in \cal{X}^*\) as additional information. 
    
    \vspace{0.1cm}
    \noindent \textbf{Ratios of anthropometric measurements}: Using the actual values of anthropometric measurements (\eg, limb lengths in \textit{mm}) from an anthropometric database, would result in a small regression error \cite{Cao_2012_16764}. We argue, though, that such information cannot be accurately obtained from state-of-the-art computer vision algorithms without employing depth information. To address this challenge, we propose to exploit the use of ratios of anthropometric measurements to alleviate the error during the estimation of the actual values. A variety of anthropometric measurements from the body and the head is provided in the CAESAR database \cite{Caesar}. We classify these measurements into two groups. The first group contains solely ratios of body measurements that can be captured from a regular surveillance camera and computed from state-of-the-art computer vision algorithms. It contains the observable features \(\bm{x}\) and will be available during both the training and the testing phases. The second group contains ratios of body measurements that are difficult to be obtained from an automated acquisition system (\eg, circumferences of body parts) and a few measurements that correspond to the head (\eg, head breadth or face length). This type of information is privileged \(\bm{x}^*\) and it will not be available at test time.
    
\begin{figure}[t] 
        \centering
        \includegraphics[width=0.4\textwidth]{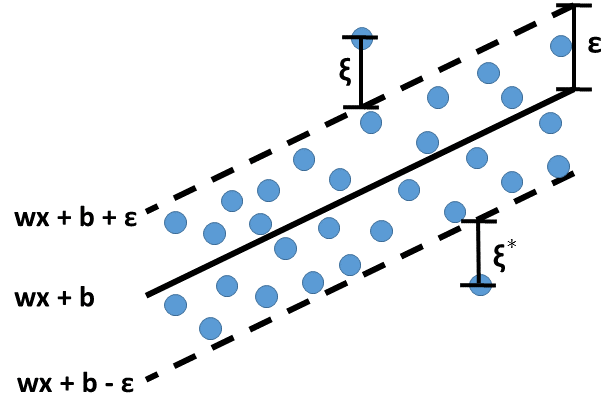}
        \caption{The $\epsilon$-insensitive band for a one-dimensional linear support vector regression problem.}
        \label{fig:SVR}
    \end{figure}

    \vspace{0.1cm}
    \noindent \textbf{$\epsilon$-SVR+}: The SVM+ method was proposed by Vapnik and Vashist \cite{Vapnik_2009_15070} to exploit privileged information for binary classification tasks. They also generalized the LUPI paradigm for the regression estimation task denoted by \(\epsilon\)-SVR. The goal in support vector regression is to find a function that has at most \(\epsilon\) deviation from the obtained targets \(y_i\) for the training set and is as flat as possible \cite{smola2004tutorial}. This means that as long as the errors are less than \(\epsilon\) they are not taken into consideration. However, any deviation larger than this will not be accepted. The standard soft-margin $\epsilon$-SVR is formulated by the following optimization problem: 
    
    where \(w \in \mathbb{R}^m\) represents the weight vector, \(||w||^2\) indicates the size of the soft margin and \(b \in\mathbb{R}\) is the bias parameter. Additionally, \(\xi_i\) is the slack variable for one training sample and indicates the deviation from the margin borders and \(C\) denotes the penalty parameter. Note that the \(\xi^*\) in $\epsilon$-SVR has nothing to do with the privileged space. It denotes the space of width \(\epsilon\) below the margin as depicted in a toy example in Figure~\ref{fig:SVR}. When privileged information is available at training time, three sets of linear functions are considered. The first set lies in the observable space in which the decision function is approximated while the other two are functions that approximate the correcting functions for the slack variables \(\xi_i\) and \(\xi_i^*\). The optimization problem is formulated as: 
    \begin{equation}\label{eq:2}
        \begin{array}{lrclcl}
            \displaystyle \underset{\substack{w,\; w_1^*,\; w_2^*\\ b,\; b_1^*,\; b_2^*}}{\text{minimize}} & \multicolumn{3}{l}{\frac{1}{2}\big(||w||^2 +\gamma(||w_1^*||^2+||w_2^*||^2)\big) + } \\
            \multicolumn{4}{l}{+ \quad C\sum_{i=1}^{l}(\langle w_1^*,x_i^*\rangle +  b_1^*) + C\sum_{i=1}^{l}( \langle w_2^*,x_i^*\rangle +  b_2^*)} \\
            \\
            \text{subject to:} & y_i - \langle w,x_i\rangle - b & \leq & \epsilon+\langle w_1^*,x_i^*\rangle +  b_1^* \\
            & \langle w,x_i\rangle + b - y_i & \leq & \epsilon+\langle w_2^*,x_i^*\rangle +  b_2^*\\
            & \langle w_1^*,x_i^*\rangle +  b_1^* & \geq & 0\\
            & \langle w_2^*,x_i^*\rangle +  b_2^* & \geq & 0\\
            & i = 1\dots l
        \end{array}
    \end{equation}
    where parameters with sub-indices equal to one and two correspond to the first and second correcting functions, respectively. Note that in both $\epsilon$-SVR and $\epsilon$-SVR+ we omit the mapping of the initial features to a higher dimensional space, because the kernel trick is used throughout our experimental investigation.
    {\renewcommand{\arraystretch}{1.1} 
    \begin{table*}[t]
        \caption{Height estimation error (\%) for a regular $\epsilon$-SVR, $\epsilon$-SVR+ and the proposed privileged information prediction (PIP) approach. Results are reported per gender and per quartile (Q).}
        \begin{center}
            \begin{tabular}{l c c c c c c c c c}
                \toprule
                & \multicolumn{3}{c}{$\epsilon$-SVR} & \multicolumn{3}{c}{$\epsilon$-SVR+} &  \multicolumn{3}{c}{\textbf{PIP}}\\
                \cmidrule(r){2-4}
                \cmidrule(r){5-7}
                \cmidrule(r){8-10}
                Q & Male & Female & Both & Male & Female & Both & Male & Female & Both \\\midrule
                \(1^{st}\)  & \(4.10 \pm 0.29\) & \(4.48 \pm 0.34\) & \(4.21 \pm 0.12\) & \(3.95 \pm 0.34\) & \(4.17 \pm 0.27\) & \(4.28 \pm 0.33\) & \(4.31 \pm 0.24\) & \(4.27 \pm 0.22\) & \(\bm{3.96 \pm 0.34}\) \\
                \(2^{nd}\)  & \(1.70 \pm 0.18\) & \(1.69 \pm 0.13\) & \(2.62 \pm 0.13\) & \(1.68 \pm 0.21\) & \(1.77 \pm 0.11\) & \(\bm{2.50 \pm 0.16}\) & \(1.80 \pm 0.12\) & \(1.71 \pm 0.19\) & \(2.65 \pm 0.12\)\\
                \(3^{rd}\)  & \(1.98 \pm 0.25\) & \(1.79 \pm 0.11\) & \(2.92 \pm 0.15\) & \(1.87 \pm 0.17\) & \(1.84 \pm 0.11\) & \(2.71 \pm 0.19\) & \(1.84 \pm 0.15\) & \(1.66 \pm 0.13\) & \(\bm{2.69 \pm 0.11}\)\\
                \(4^{th}\)  & \(4.37 \pm 0.31\) & \(4.18 \pm 0.27\) & \(4.08 \pm 0.17\) & \(4.30 \pm 0.26\) & \(4.01 \pm 0.25\) & \(3.86 \pm 0.33\) & \(3.96 \pm 0.26\) & \(3.97 \pm 0.30\) & \(\bm{3.73 \pm 0.22}\)  \\\midrule
                All &  \(3.04 \pm 0.14\) & \(3.00 \pm 0.12\) & \(3.48 \pm 0.04\)  & \(\bm{2.94 \pm 0.13}\) & \(2.91 \pm 0.12\) & \(3.33 \pm 0.10\) & \(2.95 \pm 0.12\) & \(\bm{2.89 \pm 0.10}\) & \(\bm{3.25 \pm 0.12}\) \\\bottomrule
            \end{tabular}
        \end{center}
        \label{tab:RegressionResults}
    \end{table*}
    }
    
    \begin{algorithm}[b]
        \SetKwInOut{Input}{Input}
        \SetKwInOut{Output}{Output}
        \Input{Ratios of observable \(\bm{x}\) and selected privileged \(\bm{x}^*\) features , labels \(\bm{y}\), number of selected features \(K\), \(\epsilon\), estimation error allowed \(e\)}
        \For{i= 1,...,K}{ \(//\) \textit{privileged feature prediction}\\
            \(\hat{\bm{x}}_i^* \leftarrow\) $\epsilon$-SVR model trained on \((\bm{x},\bm{x}_i^*)\)\\
        }
        \(//\) \textit{height estimation}\\
        \(h \leftarrow\) $\epsilon$-SVR model trained on \(([\bm{x}^T \;\hat{\bm{x}}^{*T}]^T,\bm{y})\)\\
        \(h_c \leftarrow\) mapping to height classes by allowing error \(e\)\\
        \Output{Height \(h\) in \textit{cm}, \(h_c \in \{1^{st}, 2^{nd}, 3^{rd}, 4^{th}\}\) quartiles}
        \caption{Privileged Information Prediction (PIP)}
        \label{alg1} 
    \end{algorithm}

    \vspace{0.1cm}
    \noindent \textbf{Privileged Information Prediction (PIP)}: We propose a novel method of estimating the height using support vector regression ($\epsilon$-SVR) in two steps. Our method takes as an input the two groups of observable and privileged human measurements and outputs its height, which is then mapped to classes (\ie, quartiles) that correspond to percentile ranges. We begin by briefly presenting the feature extraction stage because this is the first step of our pipeline. For the purposes of this paper, we assume that the anthropometric measurements of a human are available and provided to the system. However, in a real-life scenario, the observable measurements are obtained from an image of a human by applying a 3D pose estimation algorithm to obtain the location of the joints in three dimensions. The estimated skeleton is used to derive the observable measurements (\eg, arm length, hip to knee length). We also exploit another group of features (\ie, privileged measurements) such as circumferences of body parts which will be available during the training phase. Ratios of anthropometric measurements are computed for each of them, to alleviate the error that would occur during the estimation of the actual values. We then use the privileged vector \(\bm{x}^*\) to find the \(K\) most informative features denoted by \(\hat{\bm{x}}^*\) using the minimum redundancy maximum relevance feature selection (mRMR) of Peng \textit{et al.}~\cite{peng2005feature} with the mutual information difference (MID) feature selection scheme. We chose feature selection over a dimensionality reduction technique (\eg, LDA) to preserve the semantics of the features and interpret the contribution of each feature separately. For each selected feature, a support vector regression model is learned from \(\bm{x}\) that predicts its value \(\hat{\bm{x}}_i^*\). A new feature vector is formed which contains the concatenation of \(\bm{x}\) with the \(K\) predicted values of \(\hat{\bm{x}}^*\) and a new regression model is trained to predict the height. Since height is a continuous variable, performing classification (\ie, \(1^{st}\) quartile) would imply that the boundaries of the classes would have to be strictly defined, which would result in many misclassification errors. To address this challenge, we allow a percentage of error between the predicted height value and the actual height (\ie, ground truth value). Thus, if a testing sample is misclassified but the error (\%) in the estimation from the actual value is less than a threshold then we consider the sample as correctly classified. Classification accuracy results are reported in Section~\ref{sec:Exp}. 
    
    The key differences between $\epsilon$-SVR+ and the proposed approach are that the former uses information from the privileged space to add an extra term to the optimization function and further constrain the solution in the observable space. In contrast, our method employs the predictions of privileged features as extra information that can be used to estimate the height. This implies that at testing time the proposed method contains an estimation error in the feature vector that is used to predict the height. This is not the case for the $\epsilon$-SVR+ algorithm. Unlike $\epsilon$-SVR+, the proposed approach, is significantly faster to train and cross-validate despite the two regression steps instead of one. That is because the parameters that have to be tuned, except the parameter \(\epsilon\), are two for a Gaussian kernel instead of four. Finally, to the best of our knowledge, an implementation of $\epsilon$-SVR+ is not currently available, whereas the proposed approach can be re-implemented using standard programming packages.
    
    \section{Experimental Evaluation}\label{sec:Exp}
    For the purposes of this paper, we used the CAESAR database \cite{Caesar} which comprises 44 anthropometric measurements (in \textit{mm}) such as the spine-to-elbow length or the chest circumference, the weight (in \textit{kg}), and the gender of 2,392 US and Canadian civilians. After data preprocessing and discarding data with missing values, the size of the dataset we used for the experimental evaluation is 2,369 with \(39\) features for each sample, including the gender. The number of observable and privileged features are 11 and 27, respectively, whereas gender is investigated separately. Thus, the ratios of anthropometric measurements we obtain are split into: (i) \(\bm{x}\) which contains \(\nicefrac{11\times10}{2} = 55\) observable features (\ie, ratios) for each human subject and (ii) the privileged \(\bm{x}^*\) with size of \(\nicefrac{27\times26}{2} = 325\) for each sample. We opted for a Gaussian kernel for all three methods (\ie, SVR, SVR+, and PIP), which means that besides the cost parameter \(C\), the width \(\gamma_{G}\) of the kernel needs to be cross-validated. In the case of SVR+, there is an additional \(\gamma_{G}\) that needs to be cross-validated along with the parameter \(\gamma\) of the correcting space as shown in Equation~\ref{eq:2}. The possible values for all parameters were \([10^{-4}, 10^{-3}, \dots, 10^{4}] \), and a standard 5-fold cross-validation scheme was employed. Note that SVR+ requires careful selection of all four optimal parameters and thus, a full-grid search was performed. 

    Both $\epsilon$-SVR and $\epsilon$-SVR+ optimization problems are convex and can be solved using quadratic programming (QP). For large datasets and to enable fast training, a sequential minimal optimization (SMO) \cite{platt199912} technique is frequently used which divides a large QP optimization problem into a series of smaller QP problems. However, for the purposes of this paper, we opted for a regular QP solver as provided in the CVXOPT package \cite{cvxopt} and implemented the $\epsilon$-SVR+ algorithm\footnote{Our implementation can be found at \href{cbl.uh.edu/repository-code/}{\color{red}cbl.uh.edu/repository-code/}} following the dual formulation described in the work of Vapnik and Vashist \cite{Vapnik_2009_15070}. 
    
    \begin{figure}[t] 
        \centering
        \includegraphics[width=0.48\textwidth]{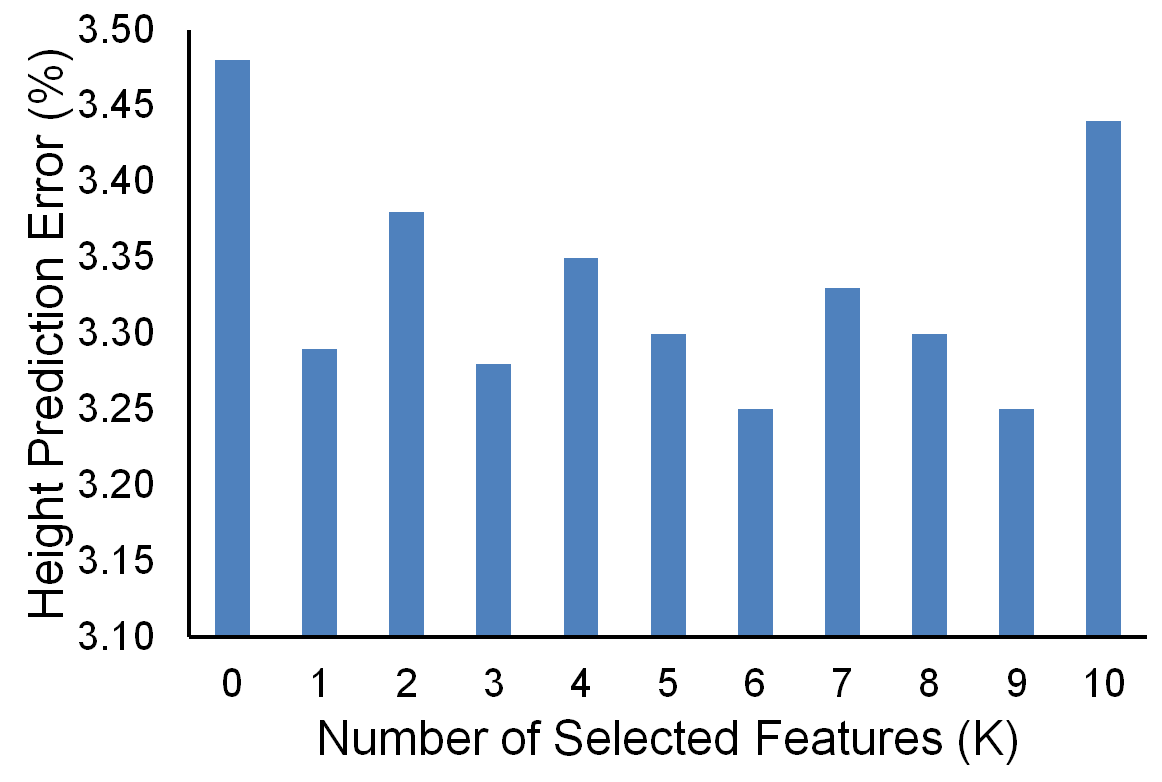}
        \caption{Height prediction error for different number of selected features.}
        \label{fig:SelectingK}
    \end{figure} 

    \vspace{0.1cm}
    \noindent \textbf{Regression-based Height Prediction}:
    We analyze and evaluate three methods: (i) a regular $\epsilon$-SVR which predicts the height attribute using ratios of observable anthropometric measurements, (ii) the $\epsilon$-SVR+ algorithm which leverages privileged information at training time to obtain a more accurate optimization solution in the observable space, and (iii) the proposed PIP approach which predicts the \(K\) most informative privileged features at test time and uses them with the observable space to estimate the height. Different models are trained and cross-validated per gender, and thus we report separate results per gender and per height quartiles (\eg, the \(1^{st}\) quartile corresponds to the shortest 25\% of the subjects). The regression error (\%) results of the aforementioned techniques are depicted in Table~\ref{tab:RegressionResults}.

    We observe that leveraging privileged information is beneficial for the estimation of soft biometric attributes such as height because both $\epsilon$-SVR+ and PIP performed better than a regular $\epsilon$-SVR. Moreover, when the gender of the human is not known beforehand, which would be the case in a real-life biometric application, the proposed approach outperformed the other two techniques in all but one case. Especially in the first and the fourth quartiles, which proved to be the most challenging ones, PIP demonstrated smaller estimation error than the other two techniques. The most challenging quartiles contain either the shortest samples (first female quartile) or the tallest subjects (fourth male quartile). The reason for this is that these groups contain heights that are close to the boundaries of the range of height values and are difficult to predict by a universal model. On the contrary, the height estimation of samples belonging to the second and the third quartiles had the smallest error in all cases. The overall performance between males and females appears to be approximately the same. Finally, using the gender of a human as prior information can reduce the height estimation error compared to a scenario in which the dataset comprises samples of both genders.  Note that a \(1\%\) error corresponds approximately to a 1.6 \textit{cm} absolute difference between the estimated and the actual value.
    
    \vspace{0.1cm}
    \noindent \textbf{Selecting the optimal number of privileged features}:
    From our investigation of predicting the privileged selected features at test time, two interesting questions arise: (i) what is considered privileged information, and (ii) what is the optimal number (\(K\)) of features to be selected that leads to best prediction accuracy? Although conceptually it might seem reasonable to use circumferences of human limbs as prior information to boost the height prediction accuracy, we observed that this is possible only through careful selection of the parameters. Second, we selected \(K\) by experimenting with different values, while concurrently performing cross-validation using the same parameters for all models to reduce the training computational time, and then estimated the height for each value. Note that our goal was to find the smallest number of \(K\), thus, from the obtained results depicted in Figure~\ref{fig:SelectingK} we set \(K = 6\). A limitation of the mRMR algorithm \cite{peng2005feature}, is that it does not consider information from groups of features but ranks them individually, which explains the fluctuations in the obtained error for a different number of selected features. 
    \begin{figure*}[t]
        \begin{subfigure}[b]{0.49\textwidth}
            \includegraphics[width=\textwidth]{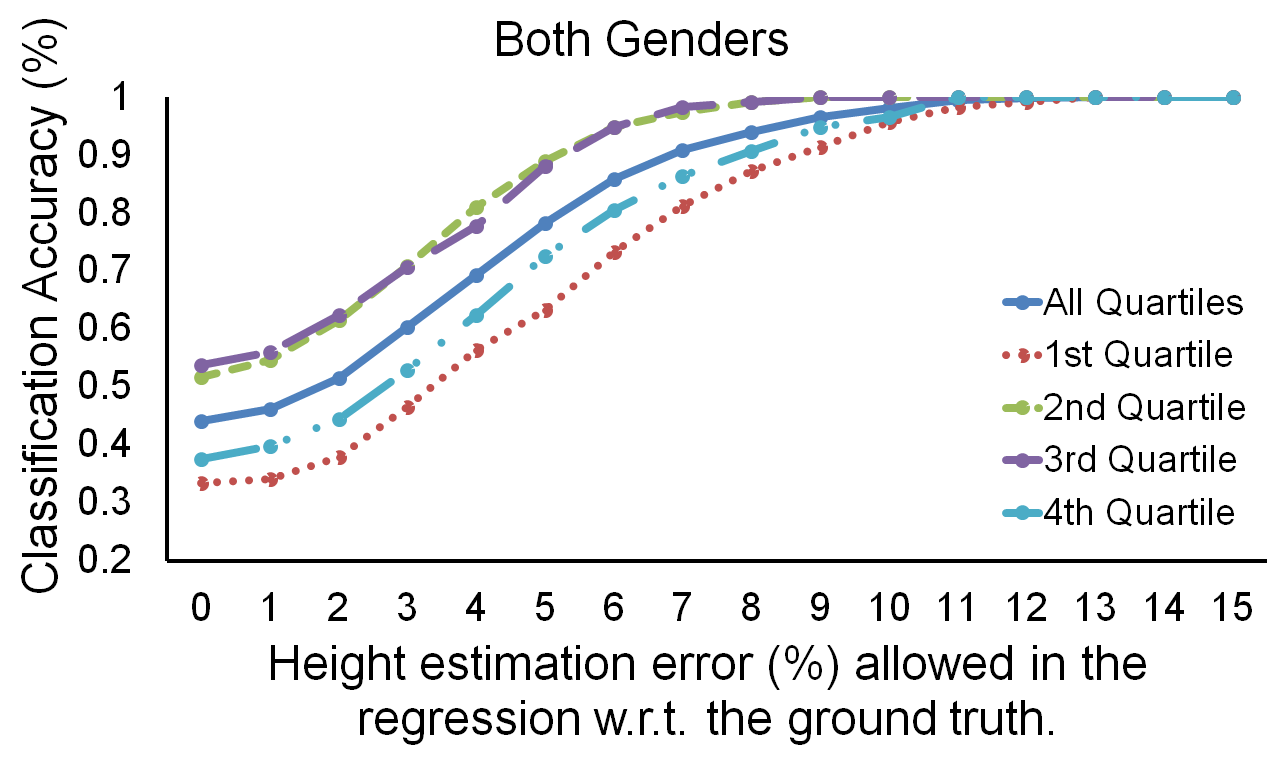}
            \caption{}
        \end{subfigure}
        \begin{subfigure}[b]{0.49\textwidth}
            \includegraphics[width=\textwidth]{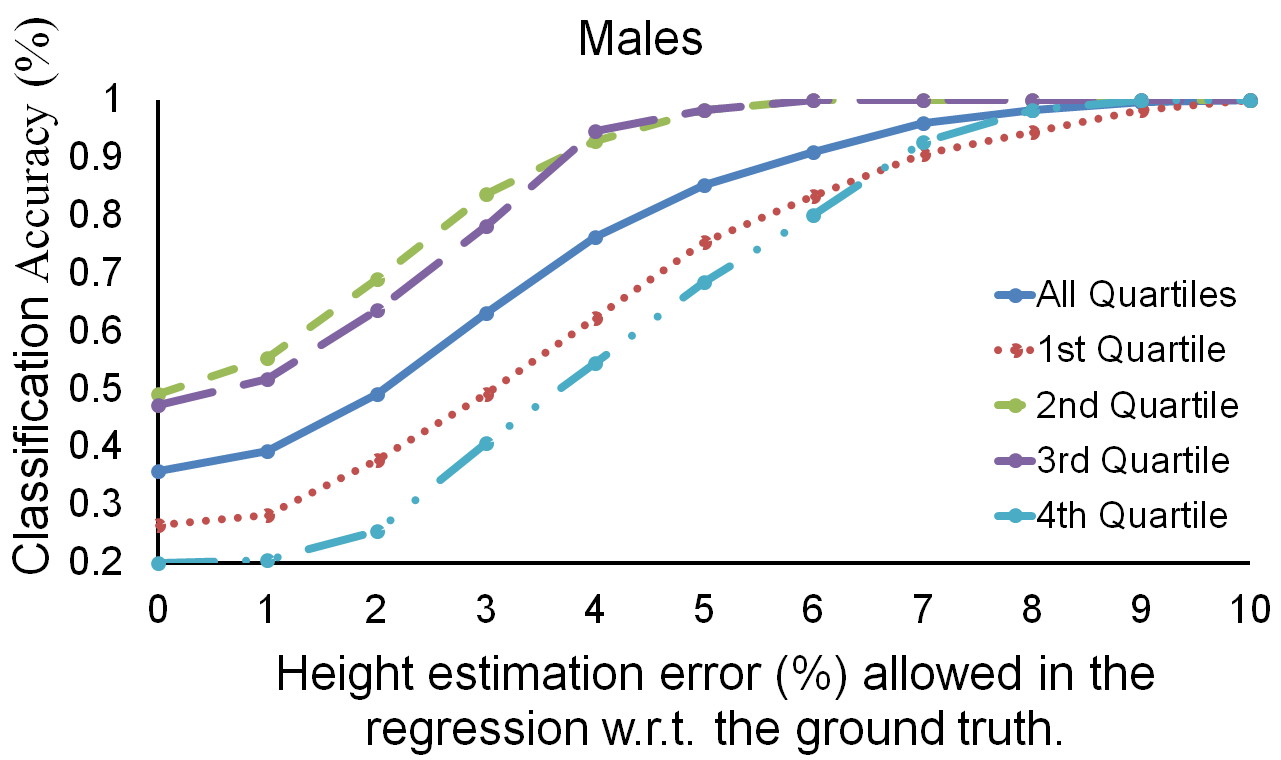}
            \caption{}
        \end{subfigure}
        \caption{Height classification results of the proposed PIP approach per quartile for (a) both genders and (b) males when the margin of error between the boundaries of the classes increases.}
        \label{fig:ClassResults}
    \end{figure*}
    
    \vspace{0.1cm}
    \noindent \textbf{Height Classification}:
    In real-life scenarios, we may seek to assign a human to a specific height class (\eg, tall male). However, height is a continuous variable, and, thus, using the \(25^{th}, 50^{th}\) and \( 75^{th}\) quantiles as the boundaries for a four-class classification problem would result in many misclassification errors. To overcome this limitation, we map the height estimation of a regular $\epsilon$-SVR to four classes (\ie, quartiles), but additionally allow a percentage of error (\(e\)) between the predicted and actual height values. This implies that if a testing sample is misclassified but the error (\%) in the estimation is less than \(e\) from the ground truth value then we consider the sample as correctly classified. The classification results obtained in this experimental investigation are shown in Figure~\ref{fig:ClassResults}. When the value of the error allowed is \(8\%\) (or 12.8 \textit{cm}) we can classify humans to four classes with an accuracy of \(90\%\) or above. Similar to the regression results, the first and the fourth quartiles have the worst classification accuracy as an error of \(13\%\) has to be allowed between the estimation and the actual value to obtain \(100\%\) classification accuracy when the gender is not known beforehand. Females are classified in the same way as males with the only difference that the quartile with the lowest classification accuracy is the first instead of the fourth. 
    
    \section{Conclusions}\label{sec:Outro}
    In this paper, we proposed a novel method for predicting privileged information at test time to perform height estimation using human metrology. Features are split to observable and privileged depending on their ability to be extracted from state-of-the-art computer vision algorithms. Ratios of anthropometric measurements are computed for every type and a feature selection scheme is employed in the privileged space to obtain the most informative features. Multiple support vector regression models are trained to predict the selected privileged features from the observable space. A new feature vector is formed from the observable features and the predictions, which is then used in a regression setup to estimate the height in \textit{cm}. A mapping to classes (\eg, \(1^{st} quartile\)) is finally performed. We observed that our approach outperforms both a regular regression scheme and the  $\epsilon$-SVR+ of Vapnik and Vashist \cite{Vapnik_2009_15070} which leverages privileged information. At the same time, it is faster to train and cross-validate. Results are reported per gender and per quartile in both regression and classification tasks, and observations are made. 
    \section*{Acknowledgments}
    This work has been funded in part by the UH Hugh Roy and Lillie Cranz Cullen Endowment Fund. The work of C. Nikou is supported by the European Commission (H2020-MSCA-IF-2014), under grant agreement No 656094. All statements of fact, opinion or conclusions contained herein are those of the authors and should not be construed as representing the official views or policies of the sponsors.
    \bibliographystyle{IEEEtran}
    \bibliography{Refs}

\begin{thebibliography}{10}
\providecommand{\url}[1]{#1}
\csname url@samestyle\endcsname
\providecommand{\newblock}{\relax}
\providecommand{\bibinfo}[2]{#2}
\providecommand{\BIBentrySTDinterwordspacing}{\spaceskip=0pt\relax}
\providecommand{\BIBentryALTinterwordstretchfactor}{4}
\providecommand{\BIBentryALTinterwordspacing}{\spaceskip=\fontdimen2\font plus
\BIBentryALTinterwordstretchfactor\fontdimen3\font minus
  \fontdimen4\font\relax}
\providecommand{\BIBforeignlanguage}[2]{{%
\expandafter\ifx\csname l@#1\endcsname\relax
\typeout{** WARNING: IEEEtran.bst: No hyphenation pattern has been}%
\typeout{** loaded for the language `#1'. Using the pattern for}%
\typeout{** the default language instead.}%
\else
\language=\csname l@#1\endcsname
\fi
#2}}
\providecommand{\BIBdecl}{\relax}
\BIBdecl

\bibitem{Jain_2004_14918}
A.~K. Jain, S.~C. Dass, and K.~Nandakumar, ``Soft biometric traits for personal
  recognition systems,'' in \emph{Proc. International Conference on Biometric
  Authentication}, Hong Kong, China, July, 15-17 2004, pp. 731--738.

\bibitem{victor2002evaluation}
B.~Victor, K.~Bowyer, and S.~Sarka, ``An evaluation of face and ear
  biometrics,'' in \emph{Proc. $16^{th}$ International Conference on Pattern
  Recognition}, vol.~1, Quebec, Canada, Aug. 11-15 2002, pp. 429--432.

\bibitem{Makinen_2008_8474}
E.~Makinen and R.~Raisamo, ``Evaluation of gender classification methods with
  automatically detected and aligned faces,'' \emph{IEEE Transactions on
  Pattern Analysis and Machine Intelligence}, vol.~30, no.~3, pp. 541--547,
  2008.

\bibitem{williams2010body}
G.~Williams, G.~Taylor, K.~Smolskiy, and C.~Bregler, ``Body motion analysis for
  multi-modal identity verification,'' in \emph{Proc. $20^{th}$ International
  Conference on Pattern Recognition}, Istanbul, Turkey, Aug. 23-26 2010, pp.
  2198--2201.

\bibitem{islam2014preliminary}
M.~R. Islam, F.~K.-S. Chan, and A.~W.-K. Kong, ``A preliminary study of lower
  leg geometry as a soft biometric trait for forensic investigation,'' in
  \emph{Proc. $22^{nd}$ International Conference on Pattern Recognition},
  Stockholm, Sweden, Aug. 24-28 2014, pp. 427--431.

\bibitem{chen2014soft}
X.~Chen and B.~Bhanu, ``Soft biometrics integrated multi-target tracking.'' in
  \emph{Proc. $22^{nd}$ International Conference on Pattern Recognition},
  Stockholm, Sweden, Aug. 24-28 2014, pp. 4146--4151.

\bibitem{D_2010_11241}
D.~Adjeroh, D.~Cao, M.~Piccirilli, and A.~Ross, ``Predictability and
  correlation in human metrology,'' in \emph{Proc. IEEE International Workshop
  on Information Forensics and Security}, Seattle, WA, Dec. 12-15 2010.

\bibitem{Caesar}
{SAE International}, ``{CAESAR}: Civilian {A}merican and {E}uropean {S}urface
  {A}nthropometry {R}esource database,'' ({A}vailable online at
  \href{http://store.sae.org.caesar}{http://store.sae.org.caesar}).

\bibitem{Cao_2012_16764}
D.~Cao, C.~Chen, D.~Adjeroh, and A.~Ross, ``Predicting gender and weight from
  human metrology using a copula model,'' in \emph{Proc. $5^{th}$ IEEE
  International Conference on Biometrics Theory, Applications and Systems},
  Washington DC, USA, Sep. 23 - 26 2012, pp. 162--169.

\bibitem{Vapnik_2009_15070}
V.~Vapnik and A.~Vashist, ``A new learning paradigm: Learning using privileged
  information.'' \emph{Neural Networks}, vol.~22, no. 5-6, pp. 544--57, 2009.

\bibitem{Sharmanska_2014_16759}
V.~Sharmanska, N.~Quadrianto, and C.~H. Lampert, ``Learning to transfer
  privileged information,'' \emph{arXiv preprint arXiv:1410.0389}, 2014.

\bibitem{Lapin_2014_16384}
M.~Lapin, M.~Hein, and B.~Schiele, ``Learning using privileged information:
  {SVM+} and weighted {SVM},'' \emph{Neural Networks}, vol.~53, pp. 95--108,
  2014.

\bibitem{Xu_2015_16245}
X.~Xu, W.~Li, and D.~Xu, ``Distance metric learning using privileged
  information for face verification and person re-identification,'' \emph{IEEE
  Transactions on Neural Networks and Learning Systems}, vol.~26, no.~12, pp.
  3150--3162, 2015.

\bibitem{Wang_2015_16756}
S.~Wang, D.~Tao, and J.~Yang, ``Relative attribute {SVM+} learning for age
  estimation,'' \emph{IEEE Transactions on Cybernetics}, vol.~46, no.~3, pp.
  827--839, 2015.

\bibitem{kakadiaris2016body}
I.~A. Kakadiaris, N.~Sarafianos, and N.~Christophoros, ``Show me your body:
  {G}ender classification from still images,'' in \emph{Proc. $23^{rd}$ IEEE
  International Conference on Image Processing}, Phoenix, AZ, Sep. 25-28 2016.

\bibitem{smola2004tutorial}
A.~J. Smola and B.~Sch{\"o}lkopf, ``A tutorial on {S}upport {V}ector
  {R}egression,'' \emph{Statistics and {C}omputing}, vol.~14, no.~3, pp.
  199--222, 2004.

\bibitem{peng2005feature}
H.~Peng, F.~Long, and C.~Ding, ``Feature selection based on mutual information
  criteria of max-dependency, max-relevance, and min-redundancy,'' \emph{IEEE
  Transactions on Pattern Analysis and Machine Intelligence}, vol.~27, no.~8,
  pp. 1226--1238, 2005.

\bibitem{platt199912}
J.~C. Platt, ``Fast training of support vector machines using sequential
  minimal optimization,'' \emph{Advances in kernel methods}, pp. 185--208,
  1999.

\bibitem{cvxopt}
M.~Andersen, J.~Dahl, and L.~Vandenberghe, ``{CVXOPT} free software package for
  convex optimization based on the {P}ython programming language. (available
  online at \href{http://cvxopt.org/}{http://cvxopt.org/}).''

\end{thebibliography}
\end{document}